\pdfoutput=1

\documentclass[11pt]{article}

\usepackage[preprint]{acl}

\usepackage{times}
\usepackage{latexsym}
\usepackage{booktabs}
\usepackage{multirow}
\usepackage{arydshln}
\usepackage{booktabs}
\usepackage{subcaption}
\usepackage{hyperref}
\usepackage{placeins}  

\usepackage[T1]{fontenc}

\usepackage[utf8]{inputenc}

\usepackage{microtype}

\usepackage{inconsolata}

\usepackage{graphicx}

\usepackage{amsmath}

\title{Fuzzy Speculative Decoding for a Tunable Accuracy-Runtime Tradeoff}

\author{Maximilian Holsman \\
  Duke University \\
  \texttt{maximilian.holsman@duke.edu} \\\And
  Yukun Huang \\
  Duke University \\
  \texttt{yukun.huang@duke.edu} \\\And
  Bhuwan Dhingra \\
  Duke University \\
  \texttt{bdhingra@cs.duke.edu}}
  
\begin{document}
\maketitle
\begin{abstract}
Speculative Decoding (SD) enforces strict distributional equivalence to the target model when accepting candidate tokens. While it maintains the target model's generation quality, this strict equivalence limits the speedup achievable by SD and prevents users from trading deviations from the target distribution in exchange for further inference speed gains. To address these limitations, we introduce \textbf{Fuzzy Speculative Decoding (FSD)} - a decoding algorithm that generalizes SD by accepting candidate tokens based on the divergences between the target and draft model distributions. By allowing for controlled divergence from the target model, FSD enables users to flexibly trade generation quality for inference speed. Across several benchmarks, our method is able to achieve significant runtime improvements of over 5 tokens per second faster than SD at only an approximate 2\% absolute reduction in benchmark accuracy. In many cases, FSD is even able to match SD benchmark accuracy at over 2 tokens per second faster, demonstrating that distributional equivalence is not necessary to maintain target model performance. Furthermore, FSD can be seamlessly integrated into existing SD extensions; we demonstrate this by applying FSD to EAGLE-2, greatly enhancing this existing extension's efficiency while allowing it to leverage FSD's tunable quality-speed tradeoff. 

\end{abstract}

\begin{figure*}[t]
\centering
  \fbox{\includegraphics[width=0.95\textwidth]{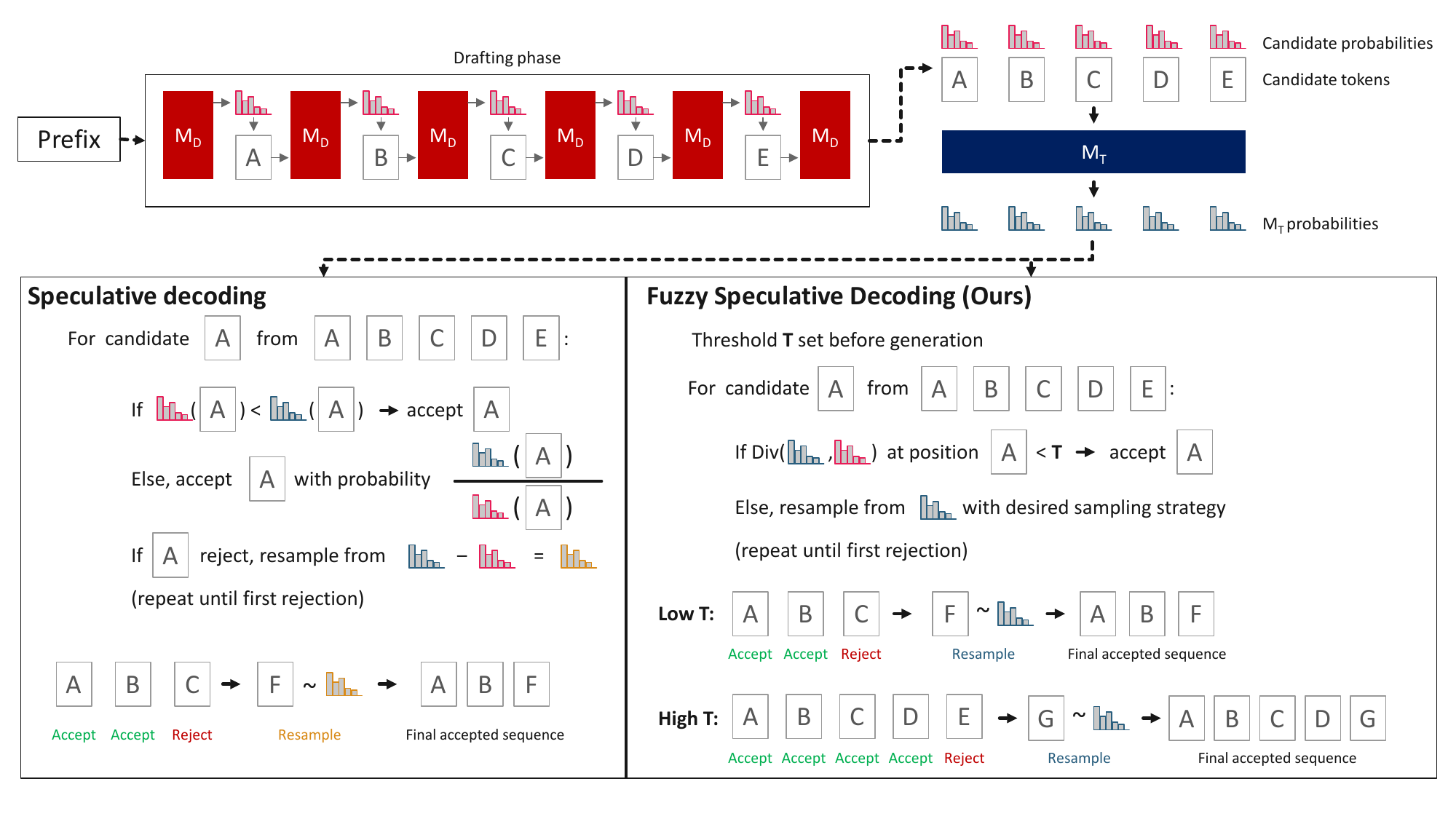}}

  \caption {Visual comparison between FSD and SD. SD accepts candidate token with a probability dependent on the candidate's relative likelihood  under $M_D$ and $M_T$. FSD determines candidate acceptance deterministically based on whether the divergence between the $M_D$ and $M_T$ distributions at the candidate's position exceeds a given threshold $T$, allowing users to determine how many candidate tokens to accept by setting the threshold $T$ accordingly.  
  }
  \label{figure:fsd_vs_sd_visual}
\end{figure*}

\section{Introduction}

Speculative decoding (SD), introduced by \citet{leviathan2023fastinferencetransformersspeculative} and \citet{chen2023acceleratinglargelanguagemodel}, is a large language model (LLM) inference acceleration algorithm that leverages a smaller, faster draft model to generate sequences of candidate tokens which are then verified and accepted in parallel by a larger target model. The speculative sampling rule that SD employs to determine which candidates to accept enforces a strict equivalence of the final sampling distribution and the original target model distribution. Thus, by cutting out the expensive sequential generation from the large target model, SD can lead to inference time reductions of around 2-3X while maintaining the same generation quality as the target model.

Despite this impressive speedup, SD suffers from two major flaws. Firstly, in order to maintain strict distributional equivalence to the target model, the SD candidate acceptance rule is overly strict, and in many cases may reject tokens that if accepted would have no impact on final generation quality \cite{lin2025criticaltokensmattertokenlevel}, thereby unnecessarily limiting potential speed-ups. Secondly, the enforced distributional equivalence means that users cannot tune the SD acceptance rule to be more or less lenient in its candidate acceptance, preventing users from trading deviations from the target model distribution in exchange for further inference speed gains. However, the flexibility for users to tune their LLM generation along an inference speed - generation quality tradeoff would be highly beneficial in real-world applications, as the relative importance of inference speed compared to generation quality may vary across different contexts within an application. 



To address these limitations of SD, we introduce \textbf{Fuzzy Speculative Decoding} - a generalized SD algorithm that determines token acceptance based on the divergence between the target and draft model distributions, allowing users to tune the generation quality - inference time tradeoff of their model. With FSD, users have the flexibility to tune a threshold parameter $T$ that determines how lenient candidate acceptance should be, and thus can control how much they are willing to deviate from the target model's distribution in exchange for further runtime reductions. As it doesn't enforce strict distributional equivalence, FSD can achieve significant runtime improvements over SD by accepting a higher percentage of candidate tokens. 

We conduct extensive experiments across four diverse benchmarks—spanning factoid QA, math, and coding—using three different model pairs. Our key findings are: 
\begin{enumerate}
    \item FSD matches SD’s accuracy while achieving over 2 tokens per second speedup by relaxing strict distributional equivalence. 
    \item FSD enables greater speedups (up to 5 tokens per second) when a slight accuracy tradeoff is acceptable (approximately 2\% absolute drop). 
    \item FSD offers a superior tunability mechanism, enabling a flexible tradeoff between accuracy and inference speed that consistently achieves higher accuracy than assigning queries between the target and draft models based on a predefined proportion.
\end{enumerate}
We also perform a broad range of ablation studies, demonstrating that FSD's performance shares many similarities with SD, including dependence on draft and target model alignment for a given text and the ability to use both sample-based and greedy decoding strategies. We also show that FSD can be applied on top of existing SD extensions like EAGLE-2, bringing the aforementioned tunability to these otherwise inflexible methods. 

\section{Previous works}
\label{section:previous_works}
Several works have sought to improve speculative decoding, primarily by increasing the acceptance rate of draft-generated tokens, including but not limited to (1) \textbf{Verifying more tokens with tree-structured proposals:}  Some methods improve efficiency by allowing the draft model to propose tokens in a tree structure, enabling the target model to verify multiple candidates in parallel using tree attention mechanisms. This expands the search space and increases the likelihood of accepting a valid token \citep{Li2024EAGLESS,  li-etal-2024-eagle, cai2024medusasimplellminference, Ankner2024HydraSD, Miao2023SpecInferAL, Chen2024SequoiaSR}. (2) \textbf{Aligning the draft model with the target model:} Methods include fine-tuning the draft model to mimic the target model’s outputs \citep{Zhou2023DistillSpecIS}, granting the draft model access to additional representation information from the target model \citep{AishwaryaP2024TandemTF, Zhang2024LearningHR, Du2024GliDeWA}, or even using a partial version of the target model as the draft model itself—such as using partial layers \citep{liu-etal-2024-speculative-decoding, elhoushi-etal-2024-layerskip, zhang-etal-2024-draft} or augmenting the target model with lightweight extensions to improve alignment \citep{Monea2023PaSSPS, Fu2024BreakTS, santilli-etal-2023-accelerating, cai2024medusasimplellminference}. (3) \textbf{Adaptive candidate length selection:} Instead of fixing the number of candidate tokens per step, some methods allow the draft model to determine when to stop generating \citep{Kim2023SpeculativeDW, Huang2024SpecDecBS}, or enable the target model to verify tokens before the draft model has finished drafting \citep{Liu2024ParallelSD}, leading to more flexible and efficient speculative decoding. While these methods enhance SD efficiency, they enforce strict distributional guarantees and offer limited flexibility in balancing accuracy and efficiency. In contrast, our framework demonstrates that such guarantees are unnecessary and provides tunable tradeoffs. Moreover, its flexibility allows seamless integration with existing approaches, paving the way for further research and optimization.

The most similar method to ours is concurrent work Judge Decoding (JD) \cite{bachmann2025judgedecodingfasterspeculative}, an SD variant where a compact module is trained on token embeddings to ‘judge’ and accept candidate tokens based on correctness rather than strict alignment with the target model. This allows JD to accept more tokens than SD with minimal performance loss. However, JD has two major limitations. First, it generalizes poorly to unseen data, as token acceptance relies on a trained judgment module. Its performance drops significantly on out-of-distribution text \cite{bachmann2025judgedecodingfasterspeculative}. \footnote{E.g., the accuracy on \texttt{HumanEval} drops from $86.6$ to $80.4\%$ when excluded from training \cite{bachmann2025judgedecodingfasterspeculative}, which would be unacceptable for most applications.} Second, JD requires per-model training, preventing out-of-the-box use for new model pairs. In contrast, FSD is training-free, generalizes across datasets, and can be applied to any model pair out-of-the-box, effectively addressing JD’s weaknesses.

\section{Speculative Decoding}

We start by reviewing how SD works in order to properly introduce FSD as an extension of this method. 

Consider a larger target model $M_T$ and a smaller draft model $M_D$. The biggest bottleneck when generating from $M_T$ individually is that tokens are sequentially dependent, and therefore each token will require a full $M_T$ forward pass to be generate conditional on the previously generated tokens. SD mitigates this bottleneck by first generating a sequence of candidate tokens sequentially from the faster $M_D$, and then uses a single $M_T$ forward pass only to \textit{verify} which of these tokens to accept. Provided that $M_D$ is a good enough approximation of $M_T$ such that a significant portion of these candidates are accepted, the runtime saved by avoiding sequential generation from $M_T$ outweighs the additional runtime of running $M_D$, resulting in an overall speedup. In order to maintain $M_T$'s full generation quality, SD accepts candidate tokens based on an acceptance rule that guarantees the final sequence of sampled tokens will still be distributed the same as they would under $M_T$. 

Specifically, at each SD step $M_D$ first generates a sequence of $L$ candidate tokens, $k=[x_0,x_1...x_L]$, which are then passed through $M_T$ to calculate the likelihood of each candidate token $x_i$ under $M_T$. Using this likelihood, each candidate $x_i$ is accepted with the probability:

$$
P_{accept}(x_i) = \min{(1, \frac{P_{M_T}(x_i|x_{<i})}{P_{M_D}(x_i|x_{<i})})}
$$
making the final candidate token SD acceptance rule:
$$
F_{accept}(x_i) = 
\begin{cases} 
    1 & \text{if } P_{accept}(x_i) > y \sim \mathcal{U}(0, 1) \\
    0, & \text{else } 
\end{cases}
$$

Once SD reaches the first rejection of the candidate sequence, it resamples a token at the rejected candidate position from the adjusted distribution: 

$$
M_{resample} = P_{M_T}(x_i|x_{<i}) - P_{M_D}(x_i|x_{<i})
$$

(Note that $P_{M_T}$ and $P_{M_D}$ will already have been calculated to determine the acceptance probability.)

By accepting tokens that are \textit{more} likely under $M_D$ than under $M_T$ with a probability of $\frac{P_{M_T}(x_i|x_{<i})}{P_{M_D}(x_i|x_{<i})}$ and resampling rejected tokens from an adjusted distribution, SD corrects for the bias introduced by $M_D$, ensuring that the final distribution remains the same as that of $M_T$. 

\begin{table}[]
\scriptsize
\centering
\resizebox{0.9\columnwidth}{!}{%
\begin{tabular}{@{}clrrr@{}}
\toprule
\textbf{Dataset} &  & \multicolumn{3}{c}{\textbf{Candidate length}} \\ \cmidrule(l){3-5} 
 &  & \textbf{5} & \textbf{10} & \textbf{15} \\ \midrule
\multirow{2}{*}{\textbf{CSQA}} & Tk. / sec & 9.3 & 9.3 & 7.9 \\  
 & \% $M_D$ Tk. & 75.7 & 82.8 & 84.6 \\ \midrule
\multirow{2}{*}{\textbf{GSM8K}} & Tk. / sec & 11.3 & 13.2 & 13.0\\  
 & \% $M_D$ Tk. & 81.5 & 89.2 & 91.4 \\ \midrule
\multirow{2}{*}{\textbf{MMLU}} & Tk. / sec & 7.2 & 7.2 & 6.3 \\  
 & \% $M_D$ Tk. & 78.7 & 85.6  & 87.5 \\ \midrule
\multirow{2}{*}{\textbf{HumanEval}} & Tk. / sec & 13.7 & 16.0 & 16.3 \\  
 & \% $M_D$ Tk. & 81.5 & 88.7 & 91.4 \\ \bottomrule
\end{tabular}%
}
\caption{Inference speeds and percent of tokens originating from $M_D$ under SD on Llama3.1 8B + 70B. Tk. / s denotes tokens per second; \% $M_D$ Tk. denotes percentage of total generated tokens originating from $M_D$.}
\label{tab:sd_tps_bp}
\end{table}

\subsection{Determining SD speed-ups}
The inference speed-up of SD heavily depends on the percentage of candidate tokens accepted. Given a fixed candidate length $L$, the more similar the distributions of $M_D$ and $M_T$ tend to be over a given generation, the more frequently candidate tokens will be accepted, and thus the greater the inference acceleration. This makes the speed-ups achieved by SD highly dependent on the distribution of text the model is generating, which we can see in Table \ref{tab:sd_tps_bp}. This variation in acceptance percentages based on text distributions means that each text  will have an optimal candidate length $L$ for which the SD inference speed is maximized. However, once the optimal $L$ has been found for the given text distribution, the percentage of tokens accepted is effectively fixed, capping the inference speed of SD to a level beyond which it cannot be increased further. This is the limitation of SD that FSD addresses.

\section{Fuzzy Speculative Decoding}

The defining difference of FSD is that it employs a different token acceptance rule that can be tuned to be more or less lenient in its acceptance decisions based on a threshold parameter $T$, which can be arbitrarily set by the user. This effectively allows users to determine how much they are willing to diverge from the target distribution $M_T$ in exchange for a higher percentage of candidates accepted, resulting in speed-ups beyond SD. 

While SD determines acceptance based on the likelihood of candidate $x_i$ under $P_{M_T}$ and $P_{M_D}$, FSD calculates the distribution-level divergence between these two distributions at each candidate position. Then, based on the tunable divergence threshold $T$, FSD will accept a candidate token if the models' divergence at the corresponding position is less than $T$. This makes the FSD acceptance rule:
$$
F_{accept}(x_i) = 
\begin{cases} 
    1 & \text{if } \text{Div}(P_{M_T}[i], P_{M_D}[i]) < T\\
    0, & \text{else } 
\end{cases}
$$

where $P_{M_T}[i]$ and $P_{M_D}[i]$ are the $M_T$ and $M_D$ next token distributions at candidate position $i$ respectively. 

In the case of candidate token rejection, FSD will sample from $P_{M_T}[i]$, that is the original target model distribution at the rejected position, with whatever sampling method the user sets for the generation. The full FSD algorithm is depicted if Figure \ref{figure:fsd_vs_sd_visual} as a side-by-side comparison with SD.

\subsection{Intuitive motivation}
Just like SD, FSD aims to accept candidate tokens at positions for which $M_T$ and $M_D$ are similar. Instead of relying on strict equivalence in final distribution, FSD relies on the fact that across an entire generation, $M_T$ and $M_D$ will produce similar tokens when the divergence between their distributions is low. This in turn means that at positions with low divergence, we can likely use tokens sampled from $M_D$ in place of those sampled from $M_T$ with minimal impact on the final generation. 

By tuning $T$, users can directly dictate how lenient candidate acceptance rule should be, thereby implicitly determining how much they are willing to allow the final sampling distribution to diverge from $M_T$ in exchange for further runtime reductions. In addition, as the FSD acceptance rule becomes more relaxed, users can also increase the candidate length $L$ past the value that was optimal for SD to realize even further reductions in inference time.

As a general framework, FSD can use any divergence type that relies solely on $P_{M_T}$ and $P_{M_D}$. In this work, we focused on KL divergence, JS divergence, and total variation distance. We define these divergences in Appendix \ref{sec:divergence_definitions}. We also perform an empirical evaluation of FSD performance under these different divergence types in Appendix \ref{appendix:div_comparison}, and find that JS divergence is the best performing divergence type for FSD. 

\subsection{Final divergence from $M_T$ under FSD}

Unlike SD, FSD does not enforce distributional equivalence to $M_T$. Tokens generated via FSD are sampled from a distribution that has diverged from $M_T$ by an amount dependent on the threshold $T$. Specifically, when generating a sequence of N tokens, the divergence between FSD sequence-level distribution and the $M_T$ sequence level distribution is upper bounded by:
$$
\text{Div}(P_{M_T}(x_{1:N}), P_{\text{FSD}}(x_{1:N})) \leq N \cdot \%_{M_D} \cdot T
$$

where  $P_{\text{FSD}}$ is the distribution of a sequence sampled from $M_D$ and $M_T$ using FSD, $N$ is the sequence length, $\%_{M_D}$ is the percentage of final tokens originating from $M_D$, and $T$ is the divergence threshold set by the user. We show the derivation of this bound in Appendix \ref{appendix:bound_derivation}.

While this bound establishes a theoretical limit on divergence, it doesn't directly indicate how FSD impacts downstream performance. The relationship between sequence-level divergence and generation quality is non-trivial, as performance degradation depends not only on the magnitude of sequence-level divergence, but also on \textit{which} tokens the models diverge on. Thus, an empirical evaluation is necessary to quantify how different choices of  $T$ impact model performance.

\subsection{Reducing FSD to standard SD}
As described, FSD accepts tokens purely based on the divergence between distributions. While we will show empirically that a sufficiently low divergence threshold can retrieve SD performance at matching or higher throughput, it may still be desirable to have FSD reduce to SD at a sufficiently low $T$. Therefore, we also introduce the following FSD variant - \textbf{reducible FSD (rFSD)} - that accepts tokens identically to SD when $T = 0$:

$$
F_{\text{accept}}(x_i) = 
\begin{cases}
1 & \begin{aligned}[t]
      & \textbf{if } \text{Div}(P_{M_T}[i], P_{M_D}[i]) < T & \\
      & \text{or } P_{\text{accept}}(x_i) > y \sim \mathcal{U}(0,1)
    \end{aligned} \\
0 & \textbf{else}\text{ resample from } M_{\text{resample}} \text{}
\end{cases}
$$


When $T=0$, $\text{Div}(P_{M_T}[i], P_{M_D}[i]) < T$ will always be false (since divergences are strictly positive for all nonequivalent distributions), reducing the acceptance rule to that of traditional SD ($P_{\text{accept}}(x_i) > y \sim \mathcal{U}(0, 1)$). In cases of rejection, rFSD would resample from the adjusted distribution $M_{resample}$ much like traditional SD would. While our main results will focus on regular FSD, we have included results in appendix \ref{appendix:rFSD_performance} showing that empirically rFSD performs the same as traditional FSD.

\section{Main experiments}

\subsection{Experiment design}
\label{section:experiment_design}

We tested FSD at various thresholds in comparison to SD on a range of benchmarks, reporting benchmark accuracy, inference speed (tokens/second), and average length of accepted candidates sequences for three of these threshold levels (denoted FSD (Low), (Med.), and (High)). We evaluated on CommonsenseQA \cite{talmor2019commonsenseqaquestionansweringchallenge} for factual knowledge, GSM8K \cite{cobbe2021trainingverifierssolvemath} for math, MMLU \cite{hendrycks2021measuringmassivemultitasklanguage} \footnote{Due to runtime constraints, we used a subset of the full MMLU dataset. This subset was sampled such that the relative prevalence of each question subject was preserved} for general knowledge and reasoning, and HumanEval \cite{chen2021evaluatinglargelanguagemodels} for code generation. We performed experiments on 3 $M_D$ - $M_T$ model pairs of varying size: Llama3.1 8B + 70B \cite{grattafiori2024llama3herdmodels}, Gemma2 2B + 27B \cite{gemmateam2024gemma2improvingopen}, and Qwen2.5 7B + 32B \cite{qwen2025qwen25technicalreport}. All Gemma2 and Qwen2.5 tests were performed on 2 A6000s, while the Llama3.1 tests were performed on 2 A100s. We use a batch size of 1 for all experiments. JS divergence was chosen as the divergence type following a preliminary experiments that indicated it performed the best. An in depth explanation of the experiment design can be found in Appendix \ref{appendix:experiment_design}, and the results of our preliminary divergence type comparison in Appendix \ref{appendix:div_comparison}. 

\subsection{Implementation}
To perform our experiments, we modified huggingface's transformers library \cite{wolf2020huggingfacestransformersstateoftheartnatural} to implemented FSD within the library's assisted generation functionality. This allows us to easily test FSD using the transformers library and allows for a fair comparison to SD, which is implemented in the library by default. We share our FSD implementation at \url{https://github.com/maxholsman/fsd}.

\subsection{FSD performance}
\begin{table*}[h]
\small
\centering
\resizebox{\textwidth}{!}{%
\begin{tabular}{llllllllllllll}
\toprule
 & \multicolumn{3}{c}{\textbf{GSM8K}} & \multicolumn{3}{c}{\textbf{CSQA}} & \multicolumn{3}{c}{\textbf{MMLU}} & \multicolumn{3}{c}{\textbf{HumanEval}} \\ 
 \midrule
 \multicolumn{14}{c}{Llama3.1 8B + 70B} \\ \midrule
 & \multicolumn{1}{c}{Acc} & \multicolumn{1}{c}{Spd} & \multicolumn{1}{c}{ALen} & \multicolumn{1}{c}{Acc} & \multicolumn{1}{c}{Spd} & \multicolumn{1}{c}{ALen} & \multicolumn{1}{c}{Acc} & \multicolumn{1}{c}{Spd} & \multicolumn{1}{c}{ALen} & \multicolumn{1}{c}{Acc} & \multicolumn{1}{c}{Spd} & \multicolumn{1}{c}{ALen} \\ \midrule

M\_D & 84.6 & 31.8 & - & 73.8 & 32.5 & - & 72.2 & 32.8 & - & 63.2 & 33.0 & - \\
M\_T & 94.9 & 8.5 & - & 83.6 & 8.9 & - & 86.2 & 9.3 & - & 79.1 & 9.3 & - \\
SD & 95.1 & 16.8 & 9.7 & 84.1 & 13.5 & 1.96 & 84.8 & 15.8 & 3.37 & 77.4 & 20.5 & 7.6 \\
FSD (Low) & 95.2 & 19.5 & 11.8 & 84.0 & 14.4 & 3.32 & 84.0 & 17.0 & 3.9 & 78.9 & 22.3 & 8.5 \\
FSD (Med.) & 94.3 & 21.2 & 12.4 & 83.7 & 17.5 & 4.3 & 83.0 & 18.1 & 4.1 & 77.6 & 23.2 & 8.6 \\
FSD (High) & 93.1 & 22.0 & 13.5 & 82.1 & 19.5 & 8.14 & 82.6 & 18.8 & 4.2 & 77.4 & 23.6 & 8.9 \\ \midrule

\multicolumn{14}{c}{Gemma2 2B + 27B} \\ \midrule
 & \multicolumn{1}{c}{Acc} & \multicolumn{1}{c}{Spd} & \multicolumn{1}{c}{ALen} & \multicolumn{1}{c}{Acc} & \multicolumn{1}{c}{Spd} & \multicolumn{1}{c}{ALen} & \multicolumn{1}{c}{Acc} & \multicolumn{1}{c}{Spd} & \multicolumn{1}{c}{ALen} & \multicolumn{1}{c}{Acc} & \multicolumn{1}{c}{Spd} & \multicolumn{1}{c}{ALen} \\ \midrule

M\_D & 57.5 & 28.5 & - & 64.6 & 31.3 & - & 55.2 & 24.3 & -  & 40.9 & 17.9 & - \\
M\_T & 90.7 & 8.8 & - & 83.0 & 9.1 & - & 75.3 & 9.4 & - & 75.6 & 9.6 & - \\
SD & 90.8 & 16.2 & 5.7 & 83.1 & 11.5 & 2.07 & 76.8 & 12.2 & 2.7 & 76.2 & 12.4 & 3.7 \\
FSD (Low) & 89.6 & 18.4 & 6.8 & 82.3 & 13.9 & 2.5 & 75.6 & 13.3 & 2.9 & 78.7 & 13.6 & 4.02 \\
FSD (Med.) & 88.5 & 19.4 & 7.1 & 81.6 & 15.7 & 3.2 & 75.4 & 15.5 & 3.5 & 77.8 & 14.1 & 4.2 \\
FSD (High) & 86.1 & 21.5 & 11.1 & 79.5 & 17.5 & 3.9 & 74.2 & 16.1 & 3.7 & 75.8 & 14.3 & 4.3 \\ \midrule

\multicolumn{14}{c}{Qwen2.5 7B + 32B} \\ \midrule
 & \multicolumn{1}{c}{Acc} & \multicolumn{1}{c}{Spd} & \multicolumn{1}{c}{ALen} & \multicolumn{1}{c}{Acc} & \multicolumn{1}{c}{Spd} & \multicolumn{1}{c}{ALen} & \multicolumn{1}{c}{Acc} & \multicolumn{1}{c}{Spd} & \multicolumn{1}{c}{ALen} & \multicolumn{1}{c}{Acc} & \multicolumn{1}{c}{Spd} & \multicolumn{1}{c}{ALen} \\ \midrule

M\_D & 89.9 & 34.8 & - & 80.2 & 36.6 & - & 71.9 & 35.6 & - & 68.1 & 26.9 & - \\
M\_T & 94.9 & 8.8 & - & 86.9 & 9.1 & - & 82.7 & 9.6 & - & 80.9 & 9.6 & - \\
SD & 95.1 & 17.4 & 6.6 & 86.8 & 14.0 & 2.7 & 82.2 & 16.0 & 3.2 & 82.1 & 15.2 & 3.7\\
FSD (Low) & 94.7 & 21.4 & 8.2 & 86.6 & 16.1 & 3.3 & 82.0 & 18.0 & 3.7 & 81.9 & 17.1 & 4.3 \\
FSD (Med.) & 94.2 & 22.4 & 9.2 & 86.1 & 19.5 & 6.6 & 81.6 & 19.5 & 4.0 & 79.0 & 17.2 & 4.4 \\
FSD (High) & 94.0 & 22.0 & 9.3 & 85.9 & 20.9 & 6.9 & 81.7 & 20.7 & 4.46 & 78.3 & 17.7 & 4.6 \\

\bottomrule
\end{tabular}
}
\caption{Benchmark performance of FSD at varying threshold levels compared to $M_D$, $M_T$, and SD. ``Acc'' refers to the QA accuracy. ``Spd'' refers to Inference Speed (tokens/sec.). ``ALen'' refers to the average accepted sequence length.} 
\label{table:main}
\end{table*}






\begin{figure*}[t]

  \includegraphics[width=\textwidth]{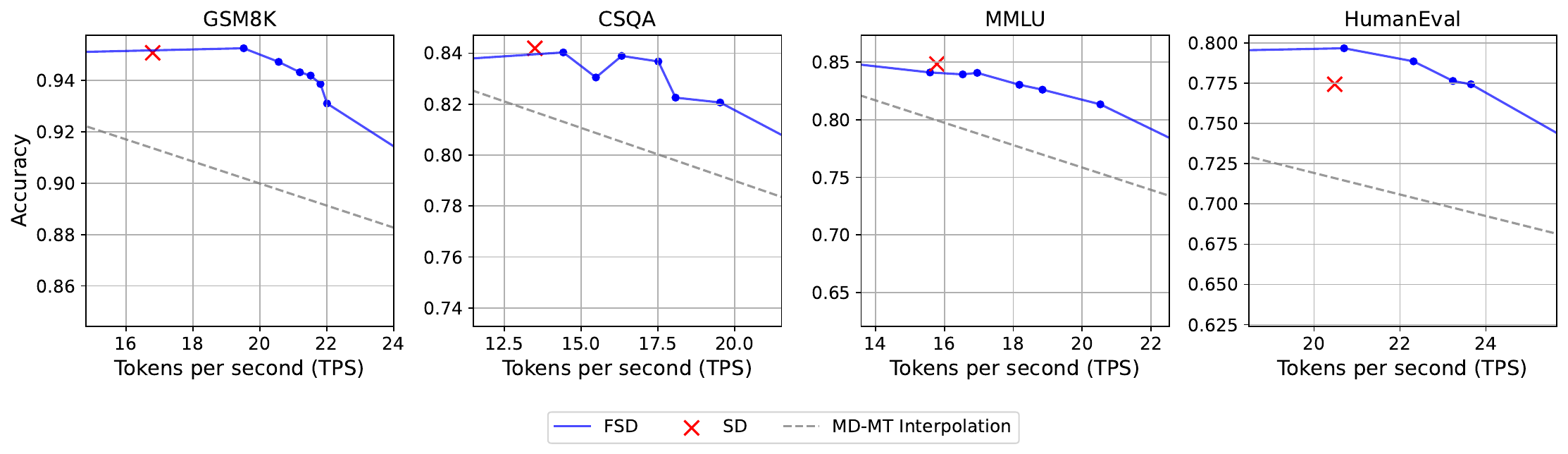}

  \caption {FSD Benchmark accuracy - inference speed tradeoff compared to SD. Results were collected with Llama3.1 8B + 70B as model pair}
  
  \label{figure:llama_main_result}
\end{figure*}

We present our experimental results in Table \ref{table:main} and in Figure \ref{figure:llama_main_result}. \\ \\ 
\textbf{FSD generally matches SD accuracy at noticeably faster inference speeds.} When setting $T$ to lower values, FSD's accuracy converges to the level of SD, often reaching this level while accepting more candidate tokens and thereby realizing greater runtime improvements. This clearly demonstrates that in many cases, the distributional equivalence enforced by SD is not necessary maintain the full $M_T$ performance level. Particularly notable are the Llama3.1 and Qwen2.5 GSM8K results, in which FSD is able to outperform SD at around 3 and 4 tokens per second faster, respectively. 

As mentioned in section \ref{section:previous_works}, many other SD extensions have been able to achieve SD performance at faster generation speeds, so this finding isn't necessarily unique to FSD. However, these prior methods all still enforce strict distributional equivalence to $M_T$, making our findings notable as they demonstrate this equivalence is often not necessary. Furthermore, given this fundamental difference, our method can easily be applied to these existing SD extensions in order to further extend their respective speedups, as we will demonstrate in section \ref{section:fsd_eagle}. \\ \\
\textbf{FSD achieves even greater runtime improvements over SD  when slight accuracy loss is acceptable.} As $T$ increases, FSD is able to achieve runtime speedups far greater than SD while only sacrificing small reductions in benchmark accuracy. The higher the divergence from $M_T$ we are willing to tolerate when accepting tokens, the greater the runtime improvement over SD. While benchmark accuracy does eventually degrade as $T$ increases, we note how minimal this deterioration is. For instance, FSD with Llama3.1 8B + 70B on CSQA achieves a 6 token per second increase over the inference speed of SD in exchange for only a 2\% absolute reduction in accuracy. We expect that in many applications of LLMs, such a runtime improvement would likely justify these small reductions in generation quality. \\ \\
\textbf{FSD allows for a previously unattainable accuracy - runtime tunability.} The accuracy - runtime tunability of FSD is demonstrated in Figure \ref{figure:llama_main_result}. A model with good tunability should satisfy two key requirements: (1) it should allow flexibility in adjusting the speed-accuracy tradeoff across the speed axis, and (2) it should achieve the highest possible accuracy compared to other methods at the same speed. Unlike SD, which has a fixed efficiency, FSD enables flexible adjustments along the speed axis while maintaining minimal accuracy degradation, thereby meeting the first requirement. To evaluate the second requirement, we introduce a \textit{random allocation} baseline, where queries are randomly assigned between the target and draft models, allowing tunability by adjusting the proportion of queries sent to the target model. We represent this baseline with a greyline interpolating between the target and draft models. As shown in Figure \ref{figure:llama_main_result}, FSD consistently outperforms the random allocation method across all speeds, demonstrating not only its flexibility but also its superior tunability.

In addition to the results above, we also share our full, more detailed results in Appendix \ref{appendix:full_results} to help characterize benchmark accuracy as a function of $T$ across the tested model pairs. 

\section{Ablation studies}

\subsection{FSD and SD variation across datasets}

As expected, the acceptance percentages and thereby the runtime improvements of both FSD and SD are highly dependent on the benchmark. We observe that FSD follows the same trends in acceptance percentages across datasets that SD does. That is, the benchmarks on which SD accept more candidate tokens (of course at the $M_T$ accuracy level) are also the benchmarks on which FSD can accepts more candidates when set to match this $M_T$ accuracy level. 

\begin{figure}[t]
  \includegraphics[width=\columnwidth]{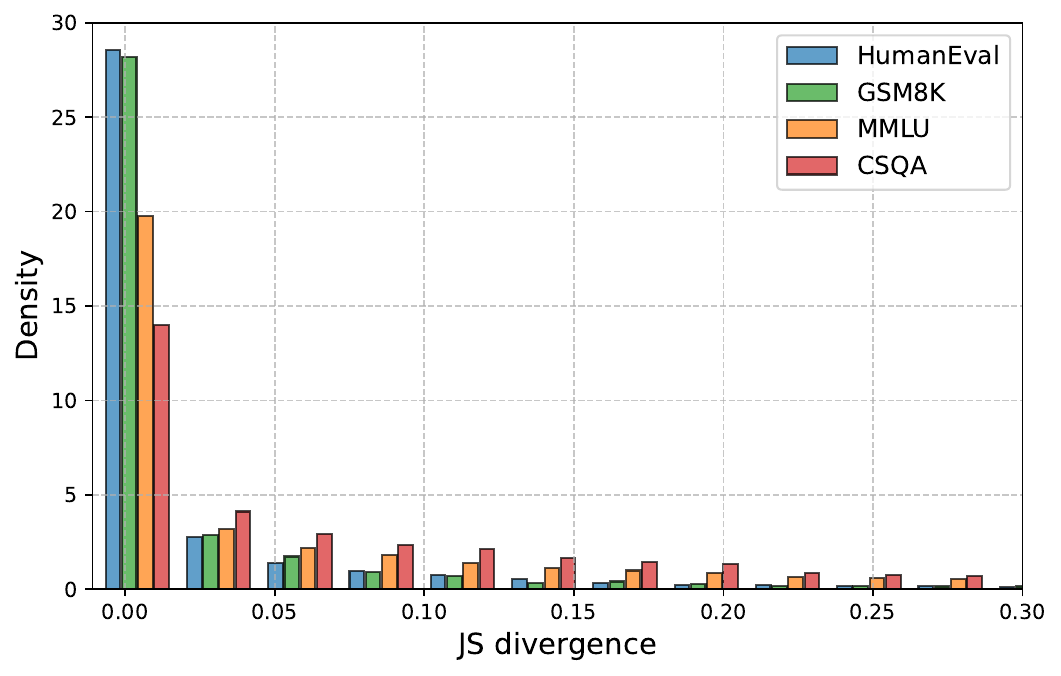}
  \caption {Distributions of JS divergences between $M_D$ and $M_T$ across tested datasets. Long tail of distributions (JS div. $\geq$ 0.3) truncated for better visibility.
  }
  \label{figure:dataset_divergence_distributions}
\end{figure}

This trend points to an underlying difference in draft and target model alignment across datasets which is affecting both methods ability to accept tokens. We illustrate this difference in $M_D$ and $M_T$ model alignment across datasets in Figure \ref{figure:dataset_divergence_distributions}, which shows the distribution of JS divergences between the Llama models on a subset of question from each dataset. As we can see, the divergences are much more heavily skewed to be much lower on datasets for which both SD and FSD accept more tokens, such as GSM8K and HumanEval. Intuitively, this makes sense: the more similar distributions tend to be across a given text generation, the lower their divergences, and thus the more candidates FSD will accept at a given threshold. Likewise, the more similar the distributions, the more likely it is that SD accepts a candidate, since the acceptance probabilities will tend to be higher. Thus, it makes sense that both FSD and SD's runtimes follow the same trend across benchmarks.

\subsection{Selecting $T$}
\label{section:selecting_t}

As previously discussed, the relationship between an FSD threshold $T$ and both the inference speed and downstream benchmark performance is dependent on the dataset and the candidate length $L$. Thus, users will not preemptively know what inference speed and downstream performance correspond to each threshold $T$ and candidate length $L$. 

However, we make two key observations that allow users to easily select T to achieve a desired performance or inference speed. First, we find that the performance level corresponding to a given threshold $T$ loosely generalizes across datasets, giving users a good starting point when setting $T$ on an unknown distribution. Second, we find that $T$ can be tuned to achieve a desired inference speed with a dev set as small as just 8-16 questions. \\ \\
\textbf{$T$-accuracy relationship across benchmarks}. Table \ref{table:fixed_fsd_threshold} shows the performance of FSD for all three model pairs at a single selected threshold held constant across datasets for each pair. We can see that for all three pairs, FSD with this constant threshold consistently achieves approximately SD accuracy at around 1-3 tokens per second faster than SD across all datasets. Thus, similar to how certain candidate lengths are known to be good starting points for SD and can later be tuned based on the specific text distribution, we show that the similar out-of-the-box values thresholds values exist for FSD. \\ \\
\textbf{Tuning T on small dev set}. We demonstrate that $T$ can be accurately tuned on a small development (dev) set in Table \ref{table:T_tuning}. For each dataset, we sampled 10 dev sets of increasing size (4 - 32 questions) from the respective training splits. Then, for a given threshold, we determine the inference speed of FSD across the dev sets, as well as on the full test set. We then calculate the average difference between the dev set inference speeds and the actual inference speed on the test set as a measure of how accurately $T$ can be tuned to achieve a given inference speed on the dev sets alone. We can see that across both datasets, even a dev set as small as just \textbf{4-8 questions} can effectively be used to tune $T$ with only minor error to "true" inference speed, demonstrating that users can easily tune $T$ to achieve a desired inference speed with very little overhead.

\begin{table*}[]
\centering
\large
\resizebox{\textwidth}{!}{%
\begin{tabular}{@{}cccc|ccc|ccc@{}}
\toprule
\multirow{2}{*}{Dataset} & \multicolumn{3}{c|}{Qwen2.5 7B + 32B} & \multicolumn{3}{c|}{Llama3.1 8B + 70B} & \multicolumn{3}{c}{Gemma2 2B + 27B} \\ \cmidrule(l){2-10} 
 & SD Acc. & \begin{tabular}[c]{@{}c@{}}FSD Acc. \\ @ T = 0.4\end{tabular} & \begin{tabular}[c]{@{}c@{}}Speedup over SD\\ (tokens / second)\end{tabular} & \multicolumn{1}{c}{SD Acc.} & \multicolumn{1}{c}{\begin{tabular}[c]{@{}c@{}}FSD Acc. \\ @ T = 0.3\end{tabular}} & \multicolumn{1}{c|}{\begin{tabular}[c]{@{}c@{}}Speedup over SD\\ (tokens / second)\end{tabular}} & \multicolumn{1}{c}{SD Acc.} & \multicolumn{1}{c}{\begin{tabular}[c]{@{}c@{}}FSD Acc. \\ @ T = 0.7\end{tabular}} & \multicolumn{1}{c}{\begin{tabular}[c]{@{}c@{}}Speedup over SD\\ (tokens / second)\end{tabular}} \\ \midrule
GSM8K & 95.1 & 94.7 & 3.4 & 95.1 & 94.7 & 3.7 & 90.8 & 89.6 & 2.1 \\ \midrule
CSQA & 86.8 & 86.4 & 3.6 & 84.2 & 83.8 & 2.8 & 83.1 & 82.3 & 2.4 \\ \midrule
MMLU & 82.3 & 82.1 & 2.8 & 84.8 & 84.1 & 1.2 & 76.8 & 74.8 & 1.9 \\ \midrule
HumanEval & 82.1 & 81.9 & 2.9 & 77.4 & 77.6 & 2.7 & 76.2 & 77.8 & 1.7 \\ \bottomrule
\end{tabular}%
}
\caption{FSD performance comparison across datasets at set thresholds}
\label{table:fixed_fsd_threshold}
\end{table*}

\begin{table}[]
\small
\resizebox{\columnwidth}{!}{%
\begin{tabular}{cccccccc}
\toprule
\multirow{2}{*}{\textbf{Dataset}} & \multirow{2}{*}{\textbf{$T$}} & \multicolumn{4}{c}{Avg. \% error w/ dev. set size n} & \multirow{2}{*}{\begin{tabular}[c]{@{}c@{}}\textbf{Test set}\\ \textbf{spd (t/s)}\end{tabular}}\\ 
\cmidrule{3-6}
 & & n = 4 & n = 8 & n = 16 & n = 32 \\
\midrule
\multirow{4}{*}{\textbf{CSQA}} 
 & 0.5 & 6.1 & 2.6 & 3.9 & 1.2 & 1.4 \\
 & 0.7 & 4.2 & 2.9 & 2.7 & 1.2 & 1.5 \\
 & 0.9 & 4.6 & 1.9 & 2.2 & 1.4 & 2.0 \\
 & 1.1 & 3.6 & 2.7 & 2.2 & 1.5 & 1.3 \\
\midrule
\multirow{4}{*}{\textbf{GSM8K}} 
 & 0.5 & 3.3 & 3.0 & 2.5 & 2.2 & 1.8 \\
 & 0.7 & 5.4 & 3.7 & 2.2 & 1.3 & 0.9 \\
 & 0.9 & 4.4 & 2.9 & 2.3 & 1.6 & 1.2 \\
 & 1.1 & 4.2 & 3.4 & 1.2 & 1.3 & 1.4 \\
\bottomrule
\end{tabular}
}
\captionof{table}{Average percentage error in inference speed between test and dev. set at increasing dev. set sizes}
\label{table:T_tuning}
\end{table}

\begin{figure}[t]
\centering
\hspace*{0.25cm}
  \includegraphics[width=0.94\columnwidth]{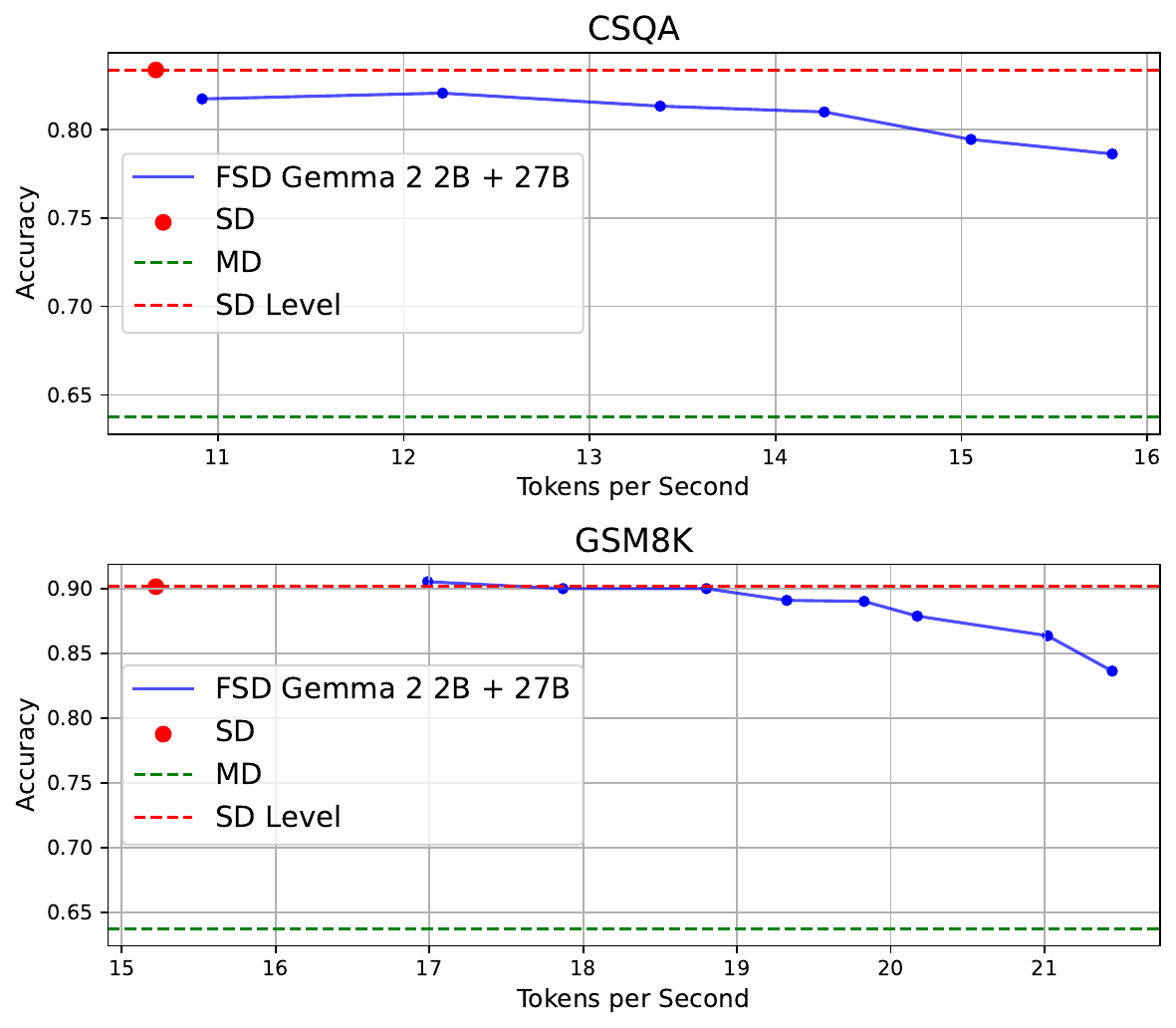}

  \caption {FSD performance compared on GSM8K and CSQA with greedy decoding from $M_T$ distribution in case of rejection. SD baselines also used greedy decoding. Model pair used was Gemma2 2B + 27B.}
  \label{figure:greedy_fsd}
\end{figure}

\begin{figure}[t]
\begin{flushright}
    \includegraphics[width=0.99\columnwidth]{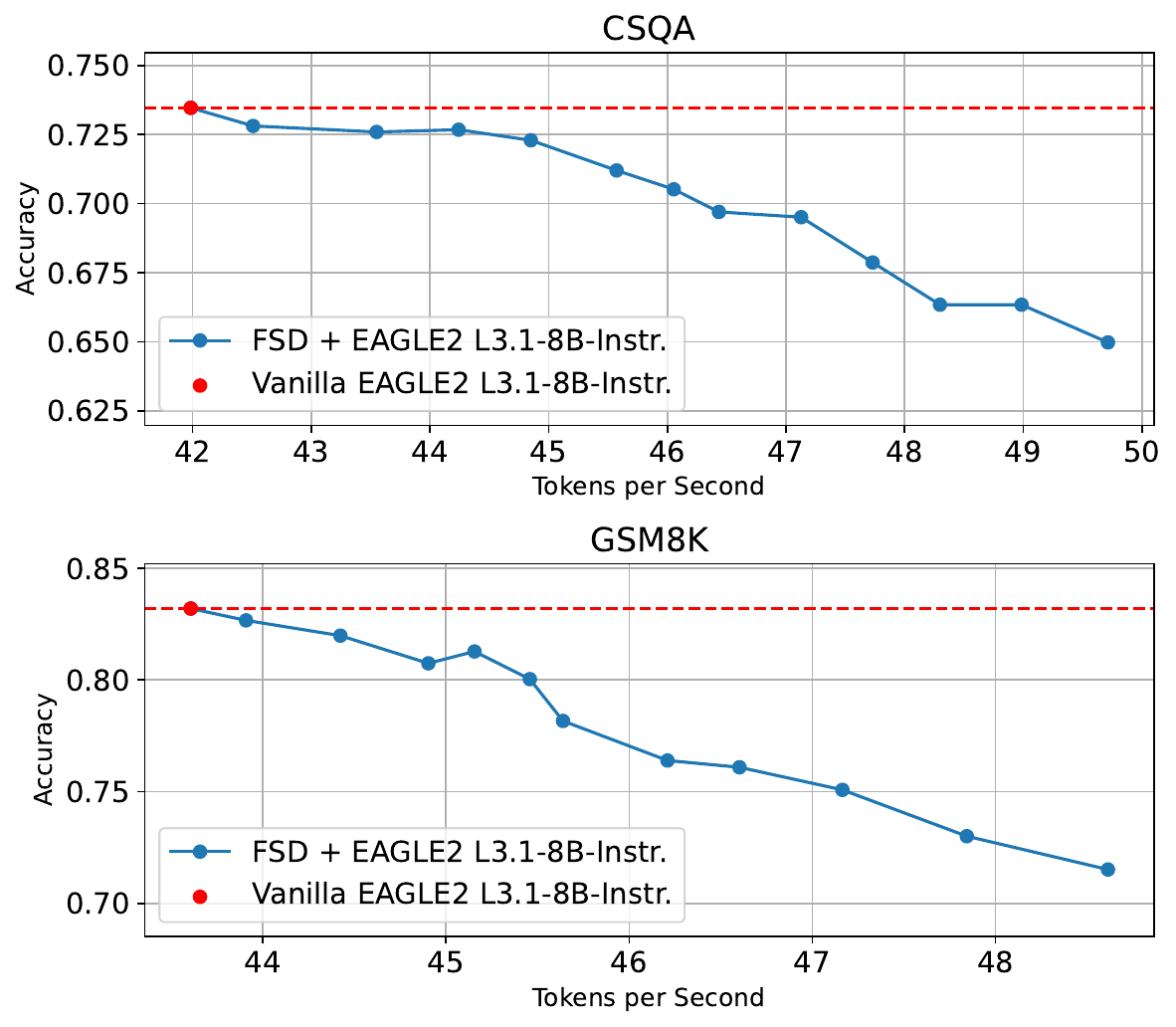}
    \caption{Accuracy-runtime tradeoff of FSD + EAGLE-2 on Llama 3.1 8B Instruct}
    \label{figure:fsd_eagle_results}
\end{flushright}
\end{figure}

\subsection{Greedy decoding vs. sample-based decoding}

As described in the experiment setup, we used greedy decoding to generate candidate sequences from $M_D$, and used sample-based decoding to sample from the $M_T$ in the case of candidate rejection. While greedy decoding from $M_D$ is standard practice when using SD, both greedy and sample-based decoding are regularly used in SD to sample from the adjusted distribution in case of rejection. Thus, the question arises whether FSD is also able to accommodate for greedy decoding, in addition to sample-based, in the case of candidate rejection.

To test this, we evaluated FSD performance on GSM8K and CSQA with greedy decoding and compared this performance to that of SD under greedy decoding, to see whether the performance trend is similar to what we observe in Table \ref{table:main}. As we can see in Figure \ref{figure:greedy_fsd}, FSD seems to follow the same performance trend observed in the main results under greedy decoding. We can again see FSD converge to SD performance at lower thresholds, and achieve significant runtime improvements at the cost of accuracy at higher thresholds. Again we can also see that the higher model alignment on GSM8K we discussed above allows FSD to achieve more impressive results over SD on this dataset, while the performance on CSQA is slightly weaker. This all is consistent with our main results in Table \ref{table:main}. 

\subsection{FSD on existing SD extensions}
\label{section:fsd_eagle}
Given its flexibility, FSD can be applied on top of most existing SD extensions, bringing tunability and even further speed-ups to the already accelerated inference that these methods achieve. Since FSD only needs the distributions of the draft and target models to determine acceptance, virtually all SD extensions can be used with FSD. To demonstrate this, we apply FSD on top of EAGLE-2 \cite{li2024eagle2fasterinferencelanguage}, and report the accuracy-runtime tradeoff on CSQA and GSM8K in Figure \ref{figure:fsd_eagle_results}. As we can see, FSD preserves the same benefits of tunability and further acceleration by foregoing distributional equivalence, even when applied to already significantly accelerated SD extensions. We expect this to hold true for most other SD extensions.

\section{Discussion}

\subsection{Potential further developments}

Unlike the probabilistic acceptance rule of SD, the FSD acceptance criteria is deterministic given the $M_D$ and $M_T$ logits. This means that FSD allows for the generation of a token-level dataset of acceptance / rejected labels, since the FSD acceptance decision relies solely on the $M_D$ and $M_T$ distributions at each tokens position. This unlocks the possibility of training a classifier to predict which tokens will be accepted and which will be rejected, based purely on the tokens up to the position being generated. Such a classifier can be used to dynamically set the candidate length generated by the draft model, reducing the number of rejected tokens at each SD step and thereby further increasing the inference time speed ups. 

The second area that we feel has potential for future development is the testing and development of a novel divergence types to identify which candidate tokens should be accepted with limited impact on generation quality. Given that FSD was already able to achieve very impressive results with simple divergence types like KL divergence and JS divergence, we expect that the divergences tailored specifically to this methods are likely to further mitigate the deterioration of generation quality as the acceptance threshold $T$ increases and allow FSD to maintain quality at even higher generation speeds. Judge Decoding \cite{bachmann2025judgedecodingfasterspeculative} attempts a similar approach to this by using learned token correctness to determine acceptance, however as discussed this method doesn't generalize, leaving this direction open for further research.

\section{Conclusion}
We have introduced FSD - a modified SD algorithm that can accept divergence allows users to tune how much divergence from $M_T$ they are willing to accept in exchange for runtime improvement beyond SD. This flexibility to achieve significantly higher runtimes, in addition an ability to match SD generation quality at faster inference in certain scenarios, makes FSD novel alternative to SD that we expect can be valuable in many LLM applications. We have shown that FSD is able to achieve very strong empirical results on-par with SD, and is able to achieve considerably higher generation speeds the cost of only minor deteriorations in generation quality. 

\section{Limitations}
The biggest limitation of our method is that it is not preemptively known what threshold $T$ will result in what downstream generation performance, as this relationship is highly dependent on the distribution of the text being generated and the candidate length $L$. Thus, as we have discussed, a practical application of FSD will have to either perform calibration tests on a text distribution similar to the eventual generation's distribution, or will have to use a potentially suboptimal out-of-the-box value similar to those discussed in section \ref{section:selecting_t}. However, we have demonstrated that tuning $T$ can be done with minimal computational overhead on a very small dev. set, and therefore does not pose a substantial barrier to the application of FSD. We also note that SD suffers from a similar reliance on hyperparameter tuning, as its inference speed is highly dependent on using the correct $L$. In fact, an incorrect selection of $L$ can result in SD having no impact on or even decreasing the generation speed compared to $M_T$. Therefore, while FSD's reliance on $T$ does represent an additional hyperparameter sensitivity, it's akin to SD’s dependency on $L$, and thereby is not a limitation unique to our method. 

\section{Acknowledgements}
Yukun Huang is supported by NSF award IIS-2211526.

\bibliography{custom, anthology}

\appendix

\section{Characterization of benchmark accuracy as function of $T$}
\label{appendix:full_results}
We share our full results with specific values of $T$ in Tables \ref{table:full_llama_results}, \ref{table:full_gemma_results}, and \ref{table:full_qwen_results}
\begin{table*}[]
\scriptsize
\centering
\resizebox{\textwidth}{!}{%
\begin{tabular}{lcccccccc}
\toprule
 & \textbf{Dataset} & \textbf{Type} & \textbf{$T$} & \textbf{Accuracy} & \textbf{Inf. Spd. (tok. / sec)} & \textbf{Acceptance length} & \textbf{Candidate length} & \textbf{Acceptance \%} \\
\midrule
\multirow{34}{*}{\textbf{Llama 3.1 8B + 70B Instruct}} & \multirow{9}{*}{\textbf{GSM8K}} & $M_T$ & - & 94.9 & 8.5 & - & - & - \\
 & & $M_D$ & - & 84.9 & 31.8 & - & - & - \\
 & & SD & - & 95.1 & 16.8 & 9.7 & 14.2 & 91.4 \\
 & & FSD & 0.3 & 95.2 & 19.5 & 11.8 & 14.2 & 92.9 \\
 & & FSD & 0.4 & 94.7 & 20.6 & 12.1 & 14.2 & 93.1 \\
 & & FSD & 0.5 & 94.3 & 21.2 & 12.4 & 14.2 & 93.3 \\
 & & FSD & 0.6 & 94.2 & 21.5 & 12.6 & 14.2 & 93.4 \\
 & & FSD & 0.7 & 93.9 & 21.8 & 13.5 & 14.2 & 93.9 \\
 & & FSD & 0.8 & 93.1 & 22.0 & 13.6 & 14.2 & 93.9 \\

\cmidrule{2-9}
 & \multirow{7}{*}{\textbf{CSQA}} 
 & $M_T$ & - & 83.6 & 8.9 & - & - & - \\
 & & $M_D$ & - & 73.8 & 32.5 & - & - & - \\
 & & SD & - & 84.2 & 13.5 & 3.0 & 4.9 & 75.4 \\
 & & FSD & 0.3 & 84.0 & 14.4 & 3.3 & 4.9 & 77.5 \\
 & & FSD & 0.35 & 83.0 & 15.5 & 3.6 & 4.9 & 79.0 \\
 & & FSD & 0.4 & 83.9 & 16.3 & 3.9 & 4.9 & 80.2 \\
 & & FSD & 0.5 & 83.7 & 17.5 & 4.3 & 4.9 & 81.8 \\
 & & FSD & 0.6 & 82.3 & 18.1 & 4.5 & 4.9 & 82.5 \\

\cmidrule{2-9}
& \multirow{8}{*}{\textbf{MMLU}} 
 & $M_T$ & - & 86.2 & 9.3 & - & - & - \\
 & & $M_D$ & - & 72.2 & 32.8 & - & - & - \\
 & & SD & - & 84.8 & 15.8 & 3.4 & 5.0 & 77.4 \\
 & & FSD & 0.3 & 84.1 & 15.6 & 3.4 & 5.0 & 77.6 \\
 & & FSD & 0.35 & 83.9 & 16.5 & 3.6 & 5.0 & 78.6 \\
 & & FSD & 0.4 & 84.1 & 17.0 & 3.8 & 5.0 & 79.6 \\
 & & FSD & 0.5 & 83.0 & 18.2 & 4.1 & 5.0 & 80.7 \\
 & & FSD & 0.6 & 82.6 & 18.8 & 4.2 & 5.0 & 81.0 \\
 & & FSD & 0.7 & 81.3 & 20.5 & 11.0 & 14.6 & 92.1 \\

\cmidrule{2-9}
 & \multirow{8}{*}{\textbf{HumanEval}} 
 & $M_T$ & - & 79.1 & 9.3 & - & - & - \\
 & & $M_D$ & - & 63.2 & 33.0 & - & - & - \\
 & & SD & - & 77.4 & 20.5 & 7.6 & 9.8 & 88.7 \\
 & & FSD & 0.2 & 79.6 & 20.7 & 8.2 & 9.8 & 89.4 \\
 & & FSD & 0.3 & 78.9 & 22.3 & 8.5 & 9.8 & 89.8 \\
 & & FSD & 0.4 & 77.6 & 23.2 & 8.8 & 9.8 & 90.1 \\
 & & FSD & 0.5 & 77.4 & 23.6 & 8.9 & 9.8 & 90.3 \\
 & & FSD & 0.6 & 75.0 & 24.8 & 9.6 & 9.8 & 90.9 \\
 & & FSD & 0.7 & 73.9 & 24.9 & 9.6 & 9.8 & 90.9 \\
\bottomrule
\end{tabular}
}
\captionof{table}{Characterized performance-accuracy tradeoff of FSD with Llama 3.1 8B + 70B Instruct}
\label{table:full_llama_results}
\end{table*}

\begin{table*}[]
\scriptsize
\centering
\resizebox{\textwidth}{!}{%
\begin{tabular}{lcccccccc}
\toprule
 & \textbf{Dataset} & \textbf{Type} & \textbf{$T$} & \textbf{Accuracy} & \textbf{Inf. Spd. (tok. / sec)} & \textbf{Acceptance length} & \textbf{Candidate length} & \textbf{Acceptance \%} \\
\midrule
\multirow{37}{*}{\textbf{Gemma 2 2B + 27B Instruct}} & \multirow{8}{*}{\textbf{GSM8K}} 
 & $M_T$ & - & 90.7 & 8.8 & - & - & - \\
 & & $M_D$ & - & 57.5 & 28.5 & - & - & - \\
 & & SD & - & 90.8 & 16.3 & 5.7 & 9.1 & 85.8 \\
 & & FSD & 0.7 & 89.6 & 18.4 & 6.8 & 9.2 & 87.8 \\
 & & FSD & 0.8 & 88.8 & 18.9 & 7.0 & 9.2 & 88.1 \\
 & & FSD & 0.9 & 88.6 & 19.4 & 7.1 & 9.2 & 88.3 \\
 & & FSD & 1.0 & 87.8 & 20.3 & 7.7 & 9.2 & 89.0 \\
 & & FSD & 1.25 & 85.9 & 21.3 & 8.1 & 9.2 & 89.6 \\
 & & FSD & 1.5 & 83.1 & 21.3 & 7.9 & 9.2 & 89.4 \\

\cmidrule{2-9}
 & \multirow{9}{*}{\textbf{CSQA}} 
 & $M_T$ & - & 83.0 & 9.1 & - & - & - \\
 & & $M_D$ & - & 64.6 & 31.3 & - & - & - \\
 & & SD & - & 83.1 & 11.5 & 2.1 & 4.7 & 68.7 \\
 & & FSD & 0.5 & 82.4 & 12.0 & 2.2 & 4.6 & 70.5 \\
 & & FSD & 0.6 & 82.1 & 13.0 & 2.4 & 4.6 & 71.7 \\
 & & FSD & 0.7 & 82.3 & 13.9 & 2.5 & 4.5 & 73.0 \\
 & & FSD & 0.8 & 81.5 & 15.1 & 3.2 & 4.5 & 77.3 \\
 & & FSD & 0.9 & 81.7 & 15.7 & 3.2 & 4.8 & 77.8 \\
 & & FSD & 1.0 & 81.1 & 16.2 & 3.3 & 4.5 & 78.5 \\
 & & FSD & 1.25 & 79.5 & 17.5 & 3.9 & 4.5 & 80.8 \\
 & & FSD & 1.5 & 77.1 & 18.0 & 3.9 & 4.5 & 81.1 \\

\cmidrule{2-9}
 & \multirow{6}{*}{\textbf{MMLU}} 
 & $M_T$ & - & 75.3 & 9.4 & - & - & - \\
 & & $M_D$ & - & 55.2 & 24.3 & - & - & - \\
 & & SD & - & 76.8 & 12.2 & 2.7 & 4.8 & 73.6 \\
 & & FSD & 0.6 & 75.6 & 13.3 & 2.9 & 4.8 & 74.9 \\
 & & FSD & 0.7 & 74.8 & 14.1 & 3.0 & 4.8 & 75.7 \\
 & & FSD & 0.8 & 75.4 & 15.5 & 3.5 & 4.8 & 78.4 \\
 & & FSD & 0.9 & 74.2 & 15.8 & 3.6 & 4.8 & 78.7 \\
 & & FSD & 1.0 & 74.2 & 16.1 & 3.7 & 4.8 & 79.1 \\

\cmidrule{2-9}
 & \multirow{8}{*}{\textbf{HumanEval}} 
 & $M_T$ & - & 75.6 & 9.6 & - & - & - \\
 & & $M_D$ & - & 40.9 & 17.9 & - & - & - \\
 & & SD & - & 76.2 & 12.4 & 3.7 & 4.9 & 78.9 \\
 & & FSD & 0.4 & 75.6 & 13.2 & 4.0 & 4.9 & 80.1 \\
 & & FSD & 0.5 & 78.7 & 13.6 & 4.0 & 4.9 & 80.3 \\
 & & FSD & 0.6 & 77.4 & 14.0 & 4.1 & 4.9 & 80.6 \\
 & & FSD & 0.7 & 77.8 & 14.1 & 4.2 & 4.9 & 80.9 \\
 & & FSD & 0.8 & 76.0 & 14.2 & 4.3 & 4.9 & 81.1 \\
 & & FSD & 0.9 & 75.8 & 14.3 & 4.3 & 4.9 & 81.4 \\
\bottomrule
\end{tabular}
}
\captionof{table}{Characterized performance-accuracy tradeoff of FSD with Gemma 2 2B + 27B Instruct}
\label{table:full_gemma_results}
\end{table*}

\begin{table*}[]
\scriptsize
\centering
\resizebox{\textwidth}{!}{%
\begin{tabular}{lcccccccc}
\toprule
 & \textbf{Dataset} & \textbf{Type} & \textbf{$T$} & \textbf{Accuracy} & \textbf{Inf. Spd. (tok. / sec)} & \textbf{Acceptance length} & \textbf{Candidate length} & \textbf{Acceptance \%} \\
\midrule
\multirow{30}{*}{\textbf{Qwen 2.5 7B + 32B Instruct}}  & \multirow{7}{*}{\textbf{GSM8K}} 
 & $M_T$ & - & 94.9 & 8.8 & - & - & - \\
 & & $M_D$ & - & 89.9 & 34.8 & - & - & - \\
 & & SD & - & 95.1 & 17.4 & 6.5 & 9.7 & 87.1 \\
 & & FSD & 0.3 & 94.8 & 20.0 & 7.8 & 9.7 & 89.1 \\
 & & FSD & 0.4 & 94.7 & 20.9 & 8.0 & 9.7 & 89.4 \\
 & & FSD & 0.5 & 94.7 & 21.4 & 8.2 & 9.7 & 89.6 \\
 & & FSD & 0.8 & 94.3 & 22.4 & 9.2 & 9.7 & 90.8 \\
 & & FSD & 0.9 & 93.9 & 22.0 & 9.3 & 9.7 & 90.8 \\

\cmidrule{2-9}
 & \multirow{6}{*}{\textbf{CSQA}} 
 & $M_T$ & - & 86.9 & 9.1 & - & - & - \\
 & & $M_D$ & - & 80.2 & 36.6 & - & - & - \\
 & & SD & - & 86.1 & 14.0 & 2.7 & 4.9 & 73.6 \\
 & & FSD & 0.4 & 86.2 & 17.6 & 6.3 & 9.7 & 87.0 \\
 & & FSD & 0.5 & 86.1 & 19.5 & 6.6 & 9.7 & 87.5 \\
 & & FSD & 0.6 & 85.9 & 20.9 & 6.9 & 9.7 & 88.1 \\

\cmidrule{2-9}
 & \multirow{6}{*}{\textbf{MMLU}} 
 & $M_T$ & - & 82.7 & 9.6 & - & - & - \\
 & & $M_D$ & - & 71.9 & 35.6 & - & - & - \\
 & & SD & - & 82.3 & 16.0 & 3.2 & 5.0 & 76.5 \\
 & & FSD & 0.3 & 82.0 & 18.0 & 3.7 & 5.0 & 79.3 \\
 & & FSD & 0.4 & 82.1 & 18.8 & 3.9 & 5.0 & 80.0 \\
 & & FSD & 0.5 & 81.6 & 19.5 & 4.0 & 5.0 & 80.6 \\
 & & FSD & 0.6 & 81.7 & 20.7 & 4.5 & 4.9 & 82.1 \\

\cmidrule{2-9}
 & \multirow{7}{*}{\textbf{HumanEval}} 
 & $M_T$ & - & 80.9 & 9.6 & - & - & - \\
 & & $M_D$ & - & 68.1 & 26.9 & - & - & - \\
 & & SD & - & 82.1 & 15.2 & 3.7 & 4.9 & 79.0 \\
 & & FSD & 0.4 & 81.9 & 17.1 & 4.3 & 4.9 & 81.2 \\
 & & FSD & 0.5 & 79.5 & 17.2 & 4.4 & 4.9 & 81.6 \\
 & & FSD & 0.6 & 79.1 & 17.4 & 4.4 & 4.9 & 81.8 \\
 & & FSD & 0.7 & 80.5 & 17.7 & 4.5 & 4.9 & 82.0 \\
 & & FSD & 0.8 & 79.9 & 17.7 & 4.5 & 4.9 & 82.1 \\
 & & FSD & 0.9 & 78.3 & 17.7 & 4.6 & 4.9 & 82.3 \\
\bottomrule
\end{tabular}
}
\captionof{table}{Characterized performance-accuracy tradeoff of FSD with Qwen 2.5 7B + 32B Instruct}
\label{table:full_qwen_results}
\end{table*}

\section{rFSD performance}
\label{appendix:rFSD_performance}
We share performance of rFSD in Table \ref{table:rFSD_performane}, demonstrating that this FSD modification performs similarly to traditional FSD.
\begin{table}[]
\small
\resizebox{\columnwidth}{!}{%
\begin{tabular}{cccccc}
\toprule
\textbf{Dataset} & \textbf{Type} & \textbf{Acc.} & \textbf{Spd. (t/s)} & \textbf{ALen} & \textbf{CLen} \\
\midrule
\multirow{4}{*}{\textbf{CSQA}} 
 & SD   & 82.9 & 7.78 & 2.06 & 5 \\
 & FSD (Low)  & 81.9 & 9.82 & 2.85 & 5 \\
 & FSD (Med.)  & 80.8 & 11.0 & 3.03 & 5 \\
 & FSD (High) & 79.6 & 11.7 & 3.22 & 5 \\
\midrule
\multirow{4}{*}{\textbf{GSM8K}} 
 & SD   & 90.0 & 10.5 & 5.72 & 10 \\
 & FSD (Low)  & 89.3 & 12.7 & 7.23 & 10 \\
 & FSD (Med.)  & 88.2 & 13.0 & 7.38 & 10 \\
 & FSD (High) & 87.0 & 13.5 & 7.56 & 10 \\
\bottomrule
\end{tabular}
}
\captionof{table}{Performance-accuracy tradeoff of rFSD compared to regular SD. Acc. indicates benchmark accuracy, Spd. indicates inference speed in tokens / second, ALen indicates acceptance length, CLen indicates candidate length}
\label{table:rFSD_performane}
\end{table}

\section{Divergence definitions}
\label{sec:divergence_definitions}

\subsection{Kullback–Leibler (KL) Divergence:}

\[
D_{\text{KL}}(P_{M_T} \| P_{M_D}) = \sum_{t \in \mathcal{V}} P_{M_T}(t \mid x) \log \left( \frac{P_{M_T}(t \mid x)}{P_{M_D}(t \mid x)} \right)
\]

where $\mathcal{V}$ is the vocabulary,  
$P_{M_T}(t \mid x)$ is the probability assigned by model $M_T$ to token $t$ given context $x$,  
$P_{M_D}(t \mid x)$ is the probability assigned by model $M_D$ to token $t$ given context $x$.

\subsection{Jensen–Shannon (JS) Divergence:}

\begin{align*}
D_{\text{JS}}(P_{M_T} \| P_{M_D}) = 
& \frac{1}{2} D_{\text{KL}}(P_{M_T} \| M) \\ 
& + \frac{1}{2} D_{\text{KL}}(P_{M_D} \| M)
\end{align*}

where $M(t \mid x)$ is the mixture distribution (average of $P_{M_T}$ and $P_{M_D}$):  
\[
M(t \mid x) = \frac{P_{M_T}(t \mid x) + P_{M_D}(t \mid x)}{2}
\]  
and $D_{\text{KL}}$ is the Kullback–Leibler divergence as defined above.

\subsection{Total Variation (TV) Distance:}

\begin{align*}
& D_{\text{TV}}(P_{M_T}, P_{M_D}) = \frac{1}{2} \sum_{t \in \mathcal{V}} \left| P_{M_T}(t \mid x) - P_{M_D}(t \mid x) \right|
\end{align*}

where:  
$P_{M_T}(t \mid x)$ and $P_{M_D}(t \mid x)$ are the probabilities from models $M_T$ and $M_D$ respectively, as defined above.

\section{Derivation of FSD sequence-level divergence bound}
\label{appendix:bound_derivation}

\subsection{KL divergence bound}

Starting with the sequence-level KL divergence decomposed autoregressively:
\begin{flalign*}
& D_{\text{KL}}(P_{M_T} \| P_{M_{FSD}}) \\
& = \sum_{t=1}^{T} E_{P_{M_T}(x_{1:t-1})} \left[ D_{\text{KL}}( P_{M_T}(t \mid x) \|  P_{M_{FSD}}(t \mid x) ) \right]
\end{flalign*}

By assumption, at each step when the $M_D$ - $M_T$ divergence exceeds $\tau$, $P_{M_T}$ is used instead of $P_{M_D}$, making the divergence 0. Let $p_{\text{use}}$ be the probability that $P_{M_D}$ is used:
\[
D_{\text{KL}}( P_{M_T}(t \mid x) \| P_{M_{FSD}}(t \mid x) ) \leq p_{\text{use}} \tau
\] \\

Summing over $T$ steps:

\begin{align*}
D_{\text{KL}}(P_{M_T} \| P_{M_D}) \leq \sum_{t=1}^{T} p_{\text{use}} \tau = T p_{\text{use}} \tau
\end{align*}

\subsection{JS divergence bound}

The JS divergence is defined as:

\begin{align*}
D_{\text{JS}}(P_{M_T} \| P_{M_D}) &= \frac{1}{2} D_{\text{KL}}(P_{M_T} \| M) \\
&\quad+ \frac{1}{2} D_{\text{KL}}(P_{M_D} \| M)
\end{align*}

Using the KL decomposition for both terms and applying the same per-step bound $\tau$ for when $P_{M_D}$ is used:
\begin{align*}
D_{\text{JS}}(P_{M_T} \| P_{M_D}) 
& \leq \frac{1}{2} \sum_{t=1}^{T} p_{\text{use}} \tau + \frac{1}{2} \sum_{t=1}^{T} p_{\text{use}} \tau \\
& = T p_{\text{use}} \tau
\end{align*}

\subsection{TV distance bound}

The sequence-level TV distance decomposes similarly via subadditivity:
\begin{align*}
& D_{\text{TV}}(P_{M_T}, P_{M_D}) \leq \\
& \sum_{t=1}^{T} E_{P_{M_T}(x_{1:t-1})} [ D_{\text{TV}}( P_{M_T}(t \mid x), P_{M_D}(t \mid x) ) ]
\end{align*}

By assumption, if $P_{M_D}$ is used, the per-step TV distance is bounded by $\tau$:
\[
D_{\text{TV}}( P_{M_T}(t \mid x), P_{M_D}(t \mid x) ) \leq p_{\text{use}} \tau
\]

Summing over $T$ steps:
\[
D_{\text{TV}}(P_{M_T}, P_{M_D}) \leq \sum_{t=1}^{T} p_{\text{use}} \tau = T p_{\text{use}} \tau
\]

\textbf{Final Result:} For all three divergences, the upper bound is:
\[
D(P_{M_T} \| P_{M_D}) \leq T p_{\text{use}} \tau.
\]
\section{Divergence comparison under FSD}
\label{appendix:div_comparison}
We referenced in the section \ref{section:experiment_design}, we performed preliminary tests on the different divergence types to see which divergence was best able to maintain SD accuracy as $T$ increases. The results of this preliminary experiments can be seen below. 

\begin{figure}[t]
  \centering
  \begin{subfigure}[t]{0.48\textwidth}
      \centering
      \includegraphics[width=\textwidth]{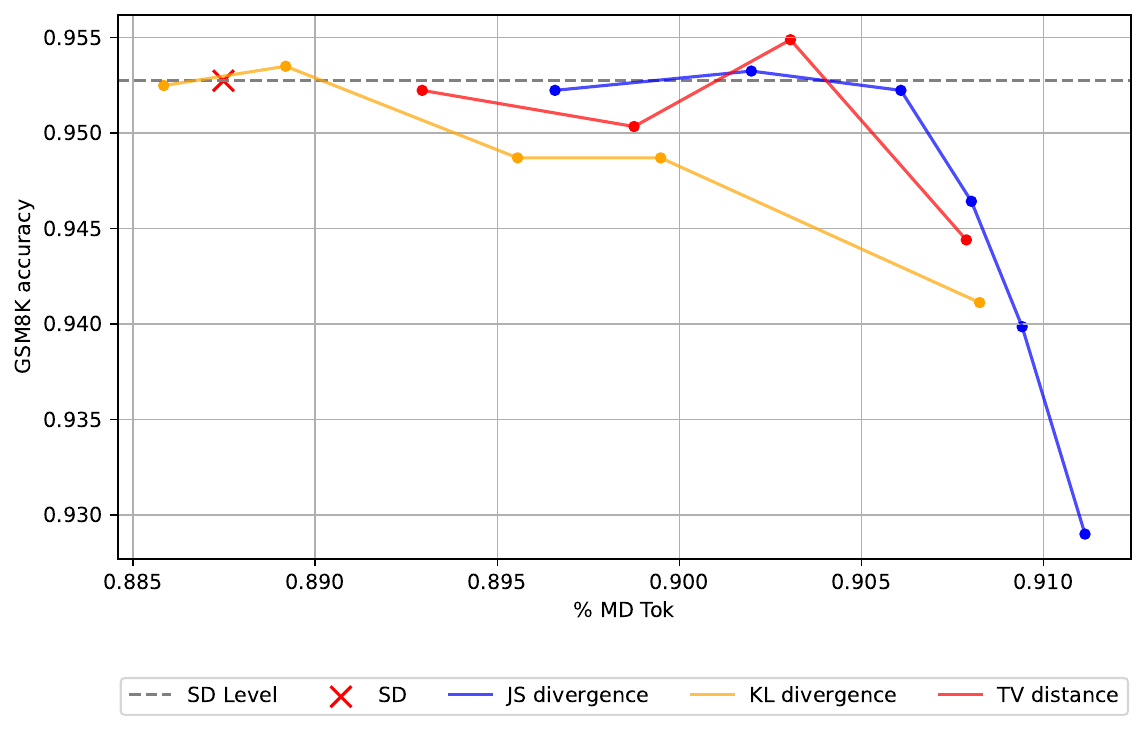}
      \caption{FSD performance of different divergence types on GSM8K.}
  \end{subfigure}%
  \hfill
  \begin{subfigure}[t]{0.48\textwidth}
      \centering
      \includegraphics[width=\textwidth]{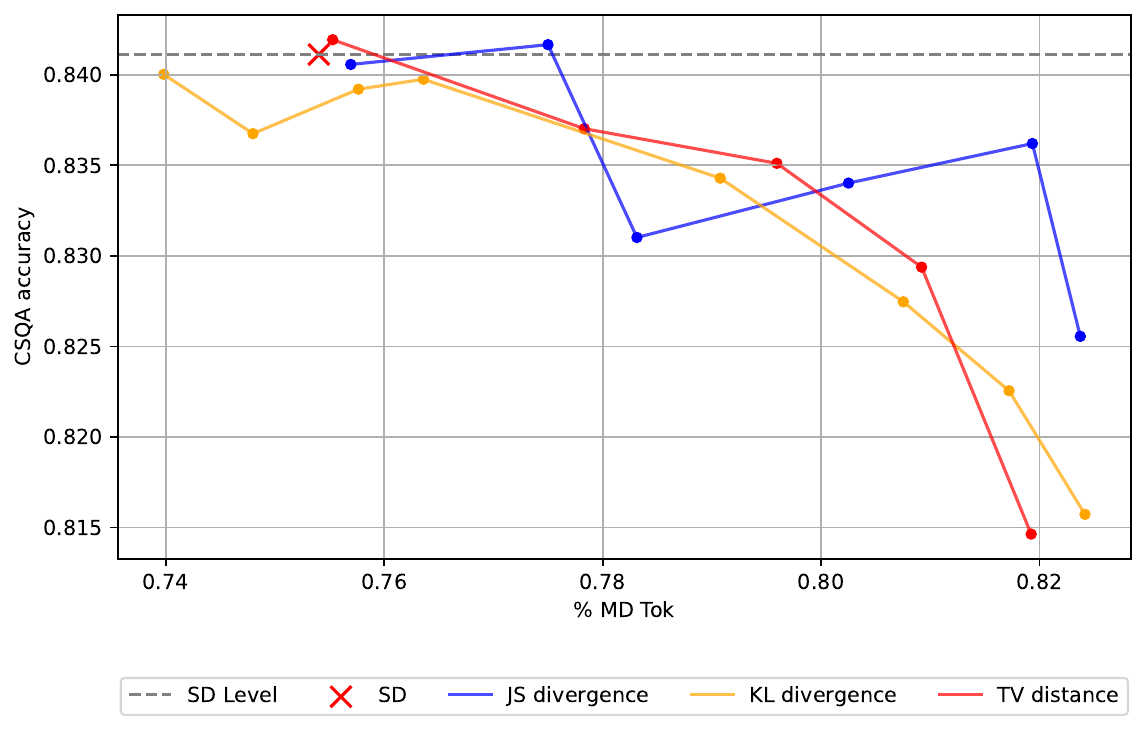}
      \caption{FSD performance of different divergence types on CSQA.}
  \end{subfigure}
  \caption{Comparison of FSD performance on GSM8K and CSQA with different divergence types.}
  \label{figure:divergence_distributions}
\end{figure}

\begin{figure}[t]
  \includegraphics[width=\columnwidth]{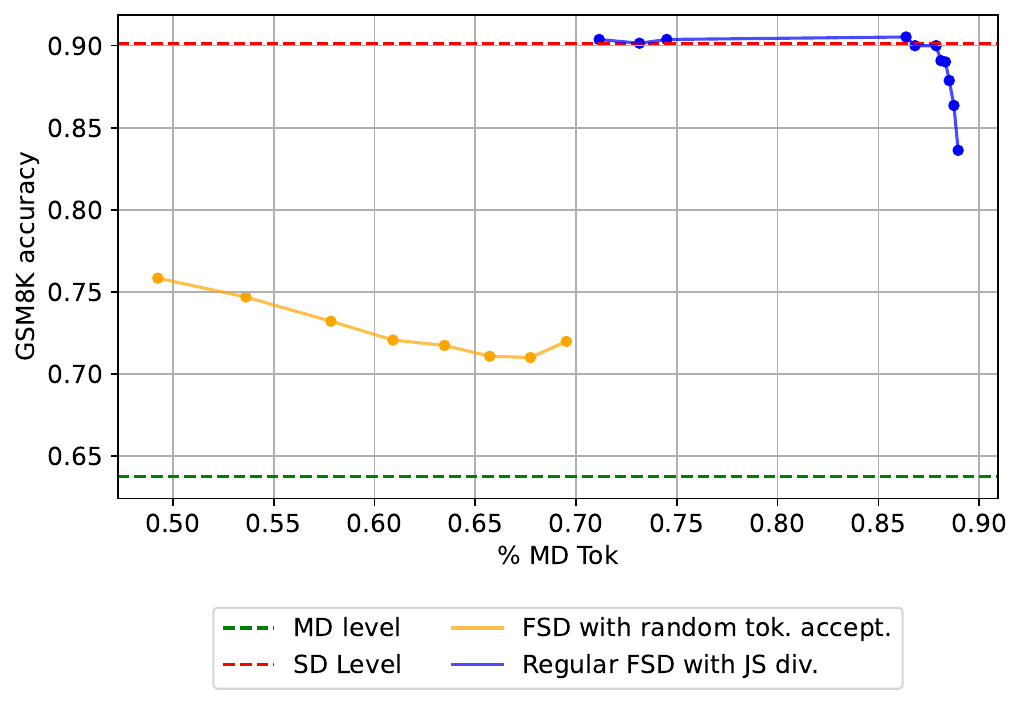}
  \caption {CSQA performance of regular FSD vs FSD with random token acceptance at varying percentages of $M_D$ tokens. \% MD Tok. denotes the percentage of final generated tokens originating from $M_D$. Experiment was performed on Gemma2 2B + 27B model pair}
  \label{figure:random_fsd}
\end{figure}

\section{In-depth experiment design}
\label{appendix:experiment_design}
Below are more details on the procedure we used to collect our main results. While this procedure was generally followed throughout, some additional data points were collected and reported (e.g., reporting $T = 0.35$ results).

For each benchmark, we start by empirically determining the approximately optimal SD candidate length $L$ by testing SD with $L = [5, 10, 15, 20]$ on a small subset of questions, and select the $L$ with the fastest inference speed as the candidate length to be used in our SD baseline. We denote this SD optimal candidate length as $L'$. We then test FSD with threshold $T=[0.1, 0.2, ... 0.9, 1.0]$ at $L'$ on the same small subset of question to determine the threshold $T_{SD}$ that accepts approximately the same percentage of candidate tokens as SD. Starting from this 'equivalent' $T_{SD}$, we then evaluate FSD's benchmark performance at threshold increasing in increments of 0.1, until benchmark performance has degraded by approximately 20\% of the performance difference between $M_D$ and $M_T$. (e.g. if $M_T$ scores 90\%, $M_D$ scores 80\%, we test FSD at increasing $T$ until accuracy reaches ~ $90 - ((90-80) * 0.2) =$ 88\%) For each threshold, we  complete three trials, using greedy decoding to generate the candidate sequences from $M_D$ and sample-based decoding to sample from $M_T$ in the case of candidate rejection. We use the same sampling strategy for our SD baseline, as this is the default for the huggingface assisted generation implementation we used.

Importantly, as the acceptance percentage increases beyond that of SD, $L'$ may no longer be the optimal candidate length. Thus, we increased $L'$ to the next highest length in $[5, 10, 15, 20]$ if we observed that FSD is accepting close to all candidates.

To quantify the performance-runtime tunability of our method, we report the FSD benchmark accuracy, inference speed (tokens/second), and average length of accepted candidates sequences at three increasing threshold levels (denoted FSD (Low), (Med.), and (High)). These three levels are meant to simulate scenarios in which users are willing to accept increasing drops in generation quality in exchange for increasing generation speeds. 

We would also like to note that a single Llama IT token was accidentally included in the Gemma2 prompt for the CSQA evaluations. We've verified that this erroneous inclusion had minimal impact on the benchmark accuracy, and have therefore retained the previous results.

\section{Random baseline}
In Table \ref{table:main}, we can clearly see that benchmark accuracy is highly sensitive to the percentage of candidate tokens accepted. For every benchmark, FSD accuracy is almost identical to SD accuracy when the threshold $T$ is set such FSD accepts a similar percentage of candidate tokens. This begs the question: is benchmark performance simply a function of the candidate acceptance percentage, irrespective of \textit{which} tokens are being accepted?

To test this, we performed a random FSD baseline, in which FSD was set to randomly accept a certain percentage of candidate tokens. By doing this, we are able to determine whether the divergences between distributions is an effective method of determining which tokens can be accepted with minimal impact on downstream performance, or whether this performance is mostly determined by \textit{how many} $M_D$ tokens are accepted. We report these results in Figure \ref{figure:random_fsd}. As expected, we can see that FSD with random candidate acceptance does significantly worse than regular divergence-based FSD, even when significantly fewer candidates from $M_D$ are being accepted. Thus, it does seem that $M_D$-$M_T$ divergence is an effective criteria for deciding which candidates to accept, implying that the development of better divergences will likely improve FSD performance even further.

\end{document}